\documentclass[10pt]{article}

\usepackage{url,color}
\usepackage{times}
\usepackage{array}
\usepackage{amsmath,amssymb}
\usepackage{graphicx}
\usepackage{amsthm}
\theoremstyle{plain}

\theoremstyle{definition}

\newcommand{\luca}{}

\title{Intrinsic Geometric Vulnerability of High-Dimensional Artificial Intelligence}

\author{Luca Bortolussi \\ Department of Mathematics and Geosciences, University of Trieste \and
Guido Sanguinetti \\ School of Informatics, University of Edinburgh}

\date{}

\begin{document}

\maketitle


\begin{abstract}
\noindent The success of modern Artificial Intelligence (AI) technologies depends critically on the ability to learn non-linear functional dependencies from large, high dimensional data sets. Despite recent high-profile successes, empirical evidence indicates that the high predictive performance is often paired with low robustness, making AI systems potentially vulnerable to adversarial attacks. In this report, we provide a simple intuitive argument suggesting that high performance and vulnerability are intrinsically coupled, and largely dependent on the geometry of typical, high-dimensional data sets. Our work highlights a major potential pitfall of modern AI systems, and suggests practical research directions to ameliorate the problem.

\end{abstract}

\section{Introduction}

Artificial Intelligence (AI) is colonising all areas of human endeavour, and its impact is widely predicted to grow exponentially in the next decades. Techniques such as deep learning have significantly improved on the state of the art in areas as diverse as computer vision, speech recognition and medical imaging  \cite{szegedy_rethinking_2015,litjens_survey_2017,simonyan_very_2014,young_recent_2017,hinton_deep_2012}, and have already reached super-human performance in games such as GO and classical ATARI video games \cite{kavukcuoglu_playing_nodate,silver_mastering_2017}. Buoyed by these successes, many researchers are heralding a new golden age for AI, and many governments and major corporations have started multi-billion dollar research investments in the development and application of AI.

Despite these undeniable achievements, the mathematical and statistical bases for AI's success, and consequently its general applicability, remain largely unclear. Techniques such as deep learning work by defining a broad class of possible input/ output functions underpinning the structure of the data. Such functional classes are encoded in the network structure, and in the so called {\it activation functions}, and are usually sufficiently general as to approximate arbitrarily well any smooth function. The specific predictive function is chosen by optimising a measure of fit to a subset of the data ({\it training data}), and performance is evaluated statistically over a held out subset of the data ({\it test set}). The training procedure (learning) is normally some variation of (stochastic) gradient descent, and much of deep learning research is concerned with the development of heuristic methods to improve the learning procedure \luca{or with the engineering of network architectures tailored to specialized tasks.} The success of this approach has largely taken by surprise even the practitioners: deep learning methods were essentially already well known in the eighties, and were largely abandoned in the intervening time as too complex and prone to overfitting.

Some attribute the new found success of deep learning methods to a combination of more powerful hardware and, crucially, much larger data sets that have become available following the advent of the internet and social networks. Recent studies on simplified models have shown how the optimisation problem itself (a notorious stumbling block for early generations of deep learning) may become simpler in the large-data regime \cite{swirszcz_local_2016,sagun_explorations_2014,choromanska_loss_2014,goodfellow_qualitatively_2014,freeman_topology_2016,baldassi2018efficiency}. However, this explanation is still unsatisfactory: it is well known that approximating a Lipschitz continuous function to a fixed precision requires a number of instances that grows exponentially with the dimension of the function's domain (e.g. \cite{traub2003information}). AI methods routinely provide excellent performance on very high-dimensional ($\sim10^4$) data sets consisting of a few million examples. These numbers may seem very large, but, in terms of learning general functions of tens of thousands of variables, they are not.\footnote{Images from the ImageNet dataset \cite{noauthor_imagenet_nodate}, typically used to train deep Convolutional Neural Network for classification, have a working resolution of 256x256 pixels, with three channels, which amounts to an input space of about $n=195,000$ dimensions. Approximating a Lipschitz function with Lipschitz constant $L$ with error $\epsilon$ requires $O((\frac{L}{\epsilon})^n)$ points, which  is a super-astronomical number even for relatively large $\epsilon$. }  

A second, less widely known limitation of deep AI methodologies is their vulnerability to adversarial attacks. As early as 2013 \cite{szegedy2013intriguing}, researchers observed that minimal perturbations to test data could completely overturn the prediction of a deep learning algorithm. For example, in a computer vision application, flipping a suitably chosen single pixel in a (correctly classified) image of a dog could return a prediction of a cat \cite{su_one_2017}. While this observation did not stop the onward march of AI (even in safety critical applications such as self-driving cars), no effective solutions to the problem of adversarial vulnerability of deep learning methods have been found. A competition held at the last  edition of the premier machine learning conference NIPS provided some promising preliminary results \cite{noauthor_nips_2017}, but unfortunately further work \cite{uesato_adversarial_2018} later showed that even these defences could be broken with a stronger attack strategy.

In this brief report, we take an alternative, geometric perspective to analyse the performance of AI methods on high-dimensional data sets. We focus on the simple case of binary classification: the prediction task consists of assigning a binary label to points in a high dimensional vector space (which we will take to be $\mathbb{R}^N$ for simplicity), based on a training set of labelled instances. Extension to multi-class classification problems is trivial. Our arguments show that indeed complex high dimensional classifiers can perform well only when the data distribution exhibit some special properties. Additionally, we show that vulnerability is a {\it direct consequence} of the structures that make learning successful, and therefore the inevitable other side of the performance medal. We focus on providing intuitive arguments that can cover the general situations, rather than rigour; proofs would be difficult to provide without strong simplifying assumptions, and would not necessarily add to our understanding of the root causes of the problem.

\section{Results}
To make progress, we start by introducing the concept of a {\it locally complex classifier}. Let $\mathcal{D}$ be a data set consisting of input/ output pairs $\{\mathbf{x}_i, y_i\}$, assumed to be drawn i.i.d. from an (unknown) distribution $p(\mathbf{x}, y)$. Input variables $\mathbf{x}$ are points in a high-dimensional vector space $\mathbf{x}\in\mathbb{R}^N$, with $N$ very large, while outputs $y$ are binary labels. A classifier is therefore a map $C\colon\mathbb{R^N}\rightarrow\{-1,1\}$ assigning to each point in input space a binary label. We will assume all classifiers to be locally constant functions, meaning that, for almost every point in input space classified as 1 (resp. -1), there exists a finite neighbourhood where the classifier does not change value. The {\it discriminant} $d_C$ defined by the classifier $C$ is  the boundary in $\mathbb{R}^N$ of the pre-image of the value 1 (or equivalently -1); from the local constancy assumption, it follows that the discriminant is a set of measure 0, and defines a surface within $\mathbb{R}^N$. The discriminant surface is the central object of study in this paper; the following definition allows us to reason precisely on the complexity of the discriminant.

\begin{figure}[!t]
\centering \includegraphics[width = 0.4\textwidth]{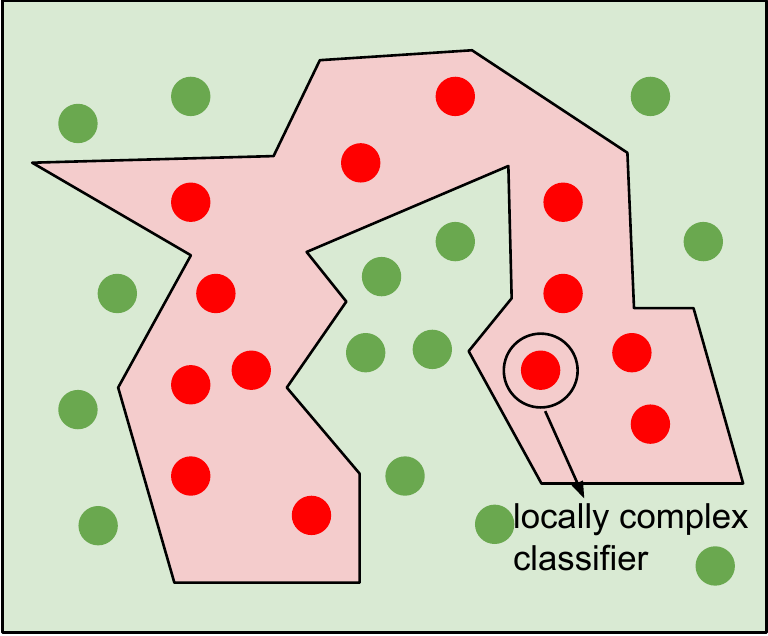}
\caption{Schematic example of locally complex discriminant} \label{fig:locally_complex}
\end{figure}

\paragraph{Definition.}
A classifier $C\colon\mathbb{R^N}\rightarrow\{-1,1\}$ is {\it locally complex} at $\mathbf{x}^*$ if the discriminant $d_C$ near $\mathbf{x}^*$ can be well approximated locally by a set of $\Omega(N)$ independent linear equations $\mathbf{w}_i^T\mathbf{x}+c=0$.
\vspace{1ex}

Intuitively, this definition captures the complexity of the discriminant by trying to quantify its "wriggliness" in high dimensions (see Figure \ref{fig:locally_complex}), requiring  the discriminant to be defined by a number of linear equalities of order $N$. Locally complex classifiers include fully grown decision trees, and deep neural networks with large numbers of nodes; in particular, deep networks using the popular rectified linear units (ReLU) activation function partition the input space in a large number of polyhedra (exponential in the number of layers) so that they are locally complex at very many points. Linear classifiers, on the other hand, express their discriminant as a single inequality, and are therefore not complex anywhere (as is to be expected).

What can the local geometry of the discriminant tell us on the performance of the classifier? Complex predictors in machine learning are often associated with overfitting problems, and indeed the following observation suggests that this problem, under certain conditions, affects all locally complex classifiers.

\paragraph{Observation.}
Let $\mathbf{x}_i$ be a training point where the classifier is locally complex, and let the (class conditional) data generating distribution be non-degenerate and with unbounded support in all directions. Then, with high probability, nearby points drawn from the data generating distribution will be misclassified.
\vspace{1ex}

This follows simply from the fact that generating a nearby point is equivalent to sampling a "noisy version" of the training point, and since the noise is unbounded in all directions the probability that in at least one of the $\Omega(N)$ ``fragile'' directions (i.e. those defining the discriminant) we sample a value that crosses the boundary grows to one exponentially in $N$. We remark that unbounded support is a very common assumption for a noise model; for example, the multivariate Gaussian distribution has support over the whole of $\mathbb{R}^N$.

This observation implies that, if a locally complex classifier performs empirically well in high dimensions, then the true data distribution must concentrate. In other words, test points must lay on a  subset of the input space of very small dimension. This observation chimes with many intuitive explanations proffered in recent years for the success of deep learning, which variously remarked on the high degree of symmetry of natural images, or on the equivalence of many local optima of the networks. In the following, we will assume that the support of the data distribution lies exactly on a low-dimensional submanifold of the ambient input space, of dimension $M \ll N$.

The true low-dimensionality of the data directly solves the conundrum of function approximation in high dimension: approximating a function requires data sets of exponentially increasing size only if the function is genuinely defined on a high dimensional domain. If all we need is a good approximation on a very small subset of the space, then the problem no longer arises. Still, the result of learning a high-dimensional classifier is a function defined on the whole ambient space. 

What will this function look like outside of the constrained data manifold? The precise answer will depend on many factors, including the training procedure and data, yet we can safely assume that it will be essentially random once sufficiently far from the data manifold. And if the data manifold is genuinely concentrated in low dimensions and embedded in a high dimensional space, sufficiently far might actually mean very near. To see why, consider the following simple example.

\begin{figure}[!t]
\centering \includegraphics[width = 0.5\textwidth]{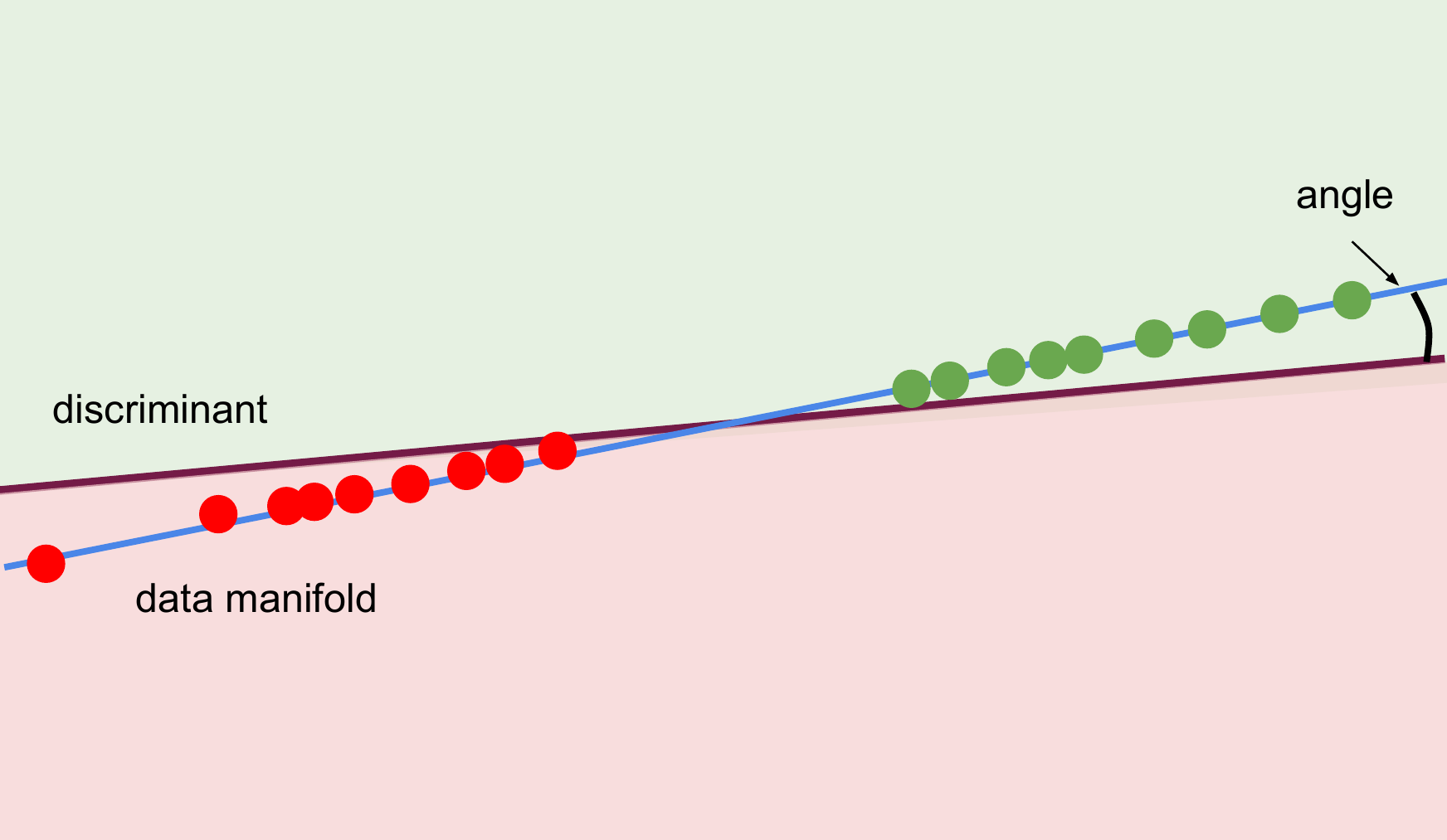}
\caption{Schematic exemplification of fragility of linear classifiers in high-dimension.} \label{fig:linear_example}
\end{figure}

\paragraph{Example: a linear classifier for apparently high dimensional data.} 
Consider a data generating distribution whose class-conditionals are well-separated Gaussians in $M\ll N$ dimensions (see Figure \ref{fig:linear_example} for N=2, M=1). Let us use logistic regression (LR) to classify this data; LR defines a hyperplane as a discriminant, and therefore requires the specification of a bias vector ($N$ parameters) and an orthogonal unit vector ($N-1$ parameters). Since the data is well separated, LR will find very accurately the optimal $M-1$ dimensional hyperplane in the data space, constraining $N+M-1$ parameters. The remaining $N-M$ parameters are unconstrained, and their value will be essentially random. If we interpret the unconstrained parameters as azimuth angles, then the distance from any data point to the discriminant will be proportional to the sine of one such angles. If $N$ is very large and $N-M$ is $O(N)$, the probability that at least one angle, hence the distance of a training point from the discriminant, will be smaller than a constant $\epsilon$ will approach 1 exponentially, therefore showing that this classifier is fragile by construction.

This property is essentially equivalent to other observations in literature about lack of robustness of linear classifiers, though it has a clearer geometric flavour. For instance, in \cite{bhagoji_enhancing_2017}, the authors observe that principal components corresponding to small eigenvalues can have an associated high weight of the linear classifier. This can be seen as the counterpart of having low angular coordinates.  This clearly explains why linear attacks are easy to find in high-dimensional models \cite{goodfellow_explaining_2014}, particularly when the data manifold has a much lower dimensionality: a step away from the manifold will typically involve a linear combination of several directions normal to the manifold having large weights, resulting in a big change in the linear classifier. Moreover, several such directions will retain large weights also when learned on a different dataset, as the weights are assigned randomly, showing that linear attacks are likely to generalise \cite{goodfellow_explaining_2014}. 

Notice that Logistic Regression is not a locally complex classifier. Indeed, under some simplifying conditions, \cite{fawzi_adversarial_2018} recently proved that {\it any} classifier in high dimensions is vulnerable when the data distribution is low dimensional. Geometry in high-dimensional spaces has also been recently advocated as a possible cause for adversarial attacks in \cite{gilmer_adversarial_2018}, where authors study a highly idealised scenario in which two-class data is distributed in two concentric spheres, and observe that misclassified points tend to appear on average close to any test point,  with a distance decreasing with the square root of the dimension.  \luca{A similar result, in a more general setting of a two classes problem on a sphere or a unit cube, has been discussed very recently in \cite{shafahi2018}, leveraging specialised isoperimetric inequalities and connecting it to some geometric properties  connected with the  the data manifold. }

\begin{figure}[!t]
\centering \includegraphics[width = 0.3\textwidth]{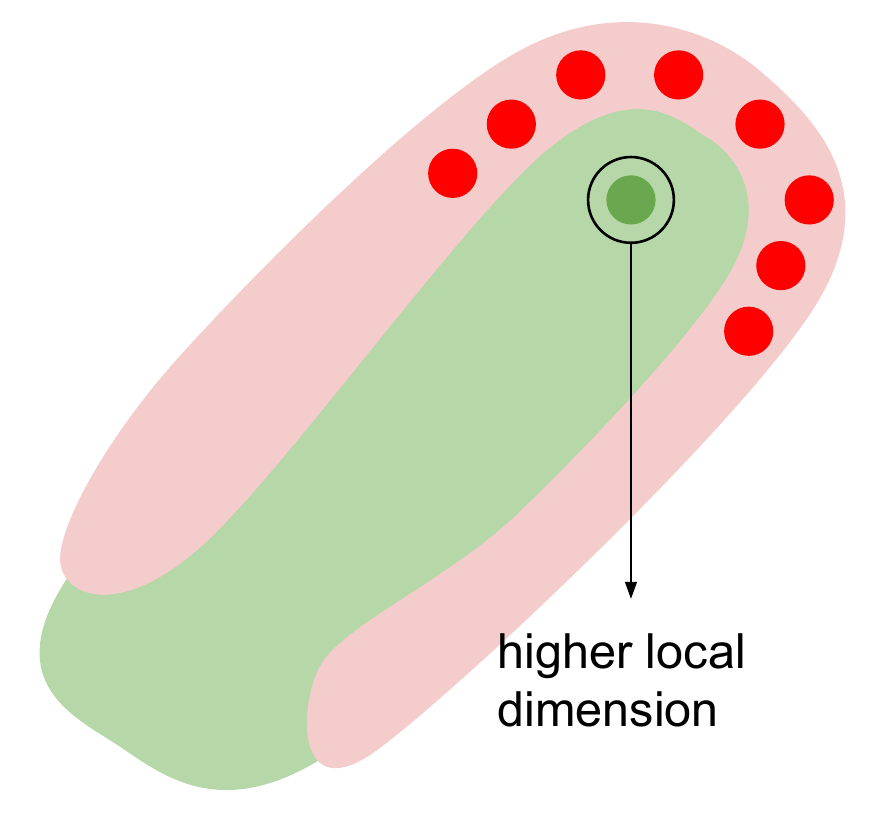}
\caption{Schematic example of increase in intrinsic local dimension.} \label{fig:local_fold}
\end{figure}

An additional intriguing feature is that many adversarial examples (i.e. small perturbations almost indistinguishable by humans that fool deep classifiers) generalise to different architectures, possibly trained on different datasets \cite{goodfellow_explaining_2014,szegedy_intriguing_2013}. Also in this case, the geometry of the data manifold and its embedding in a high dimensional space are likely to be involved. On the one hand, one can find directions that generate examples which are sufficiently far from the data manifold, the so called linear attacks in \cite{goodfellow_explaining_2014}. The high dimensionality of the input space \emph{de facto} implies that such directions exist and are common, also for simple linear models. However, not all adversarial examples are of this category, and the low dimensional data manifold itself is likely to be intrinsically complex once embedded into a high-dimensional space.  This intuitively means that each point in the data manifold is likely to be close to other parts of the manifold corresponding to different classes.  Evidence in this direction comes from the fact that adversarial examples have been found to typically have a higher local intrinsic dimensionality than training points  \cite{amsaleg_vulnerability_2017,ma_characterizing_2018}, suggesting that robust adversarial examples are found in directions in space where the data manifold folds and has a more complex local geometry (see Figure \ref{fig:local_fold}).

\section{Discussion}
In summary, our results recapitulate a number of previous observations that were broadly conjectured in the technical community, bringing them together under a novel, intuitive geometric perspective. A major new insight arising from this perspective is that complex classifiers can only work well in circumstances where they necessarily are vulnerable.

Our work also illustrates some possible directions to ameliorate the problem. Several groups are already investigating the possibility of adding local consistency constraints to the objective function of a neural network classifier \cite{szegedy_intriguing_2013,goodfellow_explaining_2014,goodfellow_generative_2014,papernot_distillation_2015,papernot_towards_2016,carlini_towards_2016} for example in the form of $\infty$-norm robustness. Such approaches show promise, yet, in order not to compromise performance, only very light regularisation can be applied. An alternative is to avoid the high-dimensionality trap by pre-processing data with a dimensionality reduction technique \cite{roweis2000nonlinear,tenenbaum2000global}. Such an approach is appealing, as it may greatly simplify the data manifold geometry, and in some simple cases can be analysed theoretically \cite{cannings2017random}, \luca{though it may still be vulnerable to white box adversarial attacks when combined in a pipeline with a (complex) classifier \cite{shafahi2018}.} Notions of intrinsic dimensionality \cite{facco_estimating_2017,ma_characterizing_2018} may play a useful role in understanding vulnerability, and indeed PCA has been advocated as a defense strategy against adversarial attacks \cite{bhagoji_enhancing_2017}.  A different direction would be to adopt a Bayesian perspective, provided we can tackle its formidable computational challenges: this would both regularise unconstrained directions and, by quantifying posterior uncertainty on model parameters, would potentially automatically detect vulnerable directions. 

\bibliographystyle{plain}
\bibliography{biblio}

\end{document}